\documentclass[10pt,twocolumn,letterpaper]{article}
\usepackage{cvpr} 
\usepackage{xspace}
\usepackage{anyfontsize}
\usepackage[dvipsnames]{xcolor}

\usepackage{amsmath}
\usepackage{amssymb}
\usepackage{amsthm}
\usepackage{amsfonts}       
\usepackage{url}            
\usepackage{booktabs}       
\usepackage{nicefrac}       
\usepackage{microtype}     
\usepackage[ruled,vlined]{algorithm2e}
\usepackage{threeparttable}
\usepackage{makecell}
\usepackage{booktabs}
\usepackage{caption}
\usepackage{wrapfig}
\usepackage{tabularx}
\usepackage{graphicx}
\usepackage{adjustbox}
\usepackage{bm}
\usepackage{tablefootnote}
\usepackage{multirow}
\usepackage{pifont}
\usepackage{array}
\usepackage{subcaption}
\usepackage{booktabs}
\usepackage{varwidth}
\usepackage{nicematrix,tikz}
\usepackage{amsmath}
\usepackage{verbatim}
\usepackage{siunitx}

\NiceMatrixOptions
  {
    custom-line = 
     {
       letter = : ,
       command = dashedline , 
       ccommand = cdashedline ,
       tikz = dashed
     }
  }
  
\newcommand{\Tref}[1]{Table~\ref{#1}}
\newcommand{\eref}[1]{Eq.~\eqref{#1}}

\newcommand{\fref}[1]{Fig.~\ref{#1}}

\newcounter{todos}
\AtEndDocument{\ifnum\value{todos}>0 \PackageWarning{TODOS}{There are \arabic{todos} todos left in this paper! Fix them before submitting the paper!} \fi}
\newcommand{\todo}[1]{\protect\stepcounter{todos}{\color{red}{[TODO: #1]}}}
\newcommand{\fl}[1]{\protect\stepcounter{todos}{\color{magenta}{[feiran-l: #1]}}}

\makeatletter
\DeclareRobustCommand\onedot{\futurelet\@let@token\@onedot}
\def\@onedot{\ifx\@let@token.\else.\null\fi\xspace}

\def\wrt{w.r.t\onedot} 

\makeatother

\definecolor{cvprblue}{rgb}{0.21,0.49,0.74}
\usepackage[pagebackref,breaklinks,colorlinks,allcolors=cvprblue]{hyperref}

\def\paperID{7498} 
\def\confName{CVPR}
\def\confYear{2026}

\title{PAGen: Phase-guided Amplitude Generation for Domain-adaptive\\ Object Detection}

\author{Shuchen Du
\quad
Shuo Lei
\quad
Feiran Li
\quad
Jiacheng Li
\quad
Daisuke Iso
\\
Sony Research
\\
{\tt\small \{shuchen.du, shuo.lei, feiran.li, jiacheng.li, daisuke.iso\}@sony.com}
}

\begin{document}
\maketitle
\begin{abstract}

Unsupervised domain adaptation (UDA) greatly facilitates the deployment of neural networks across diverse environments. However, most state-of-the-art approaches are overly complex, relying on challenging adversarial training strategies, or on elaborate architectural designs with auxiliary models for feature distillation and pseudo-label generation. In this work, we present a simple yet effective UDA method that learns to adapt image styles in the frequency domain to reduce the discrepancy between source and target domains. The proposed approach introduces only a lightweight pre-processing module during training and entirely discards it at inference time, thus incurring no additional computational overhead. We validate our method on domain-adaptive object detection (DAOD) tasks, where ground-truth annotations are easily accessible in source domains (e.g., normal-weather or synthetic conditions) but challenging to obtain in target domains (e.g., adverse weather or low-light scenes). Extensive experiments demonstrate that our method achieves substantial performance gains on multiple benchmarks, highlighting its practicality and effectiveness.

\end{abstract}    
\section{Introduction}
\label{sec:intro}

\begin{figure}[t]
  \centering
  \hfill
  \includegraphics[width=\linewidth,height=6cm]{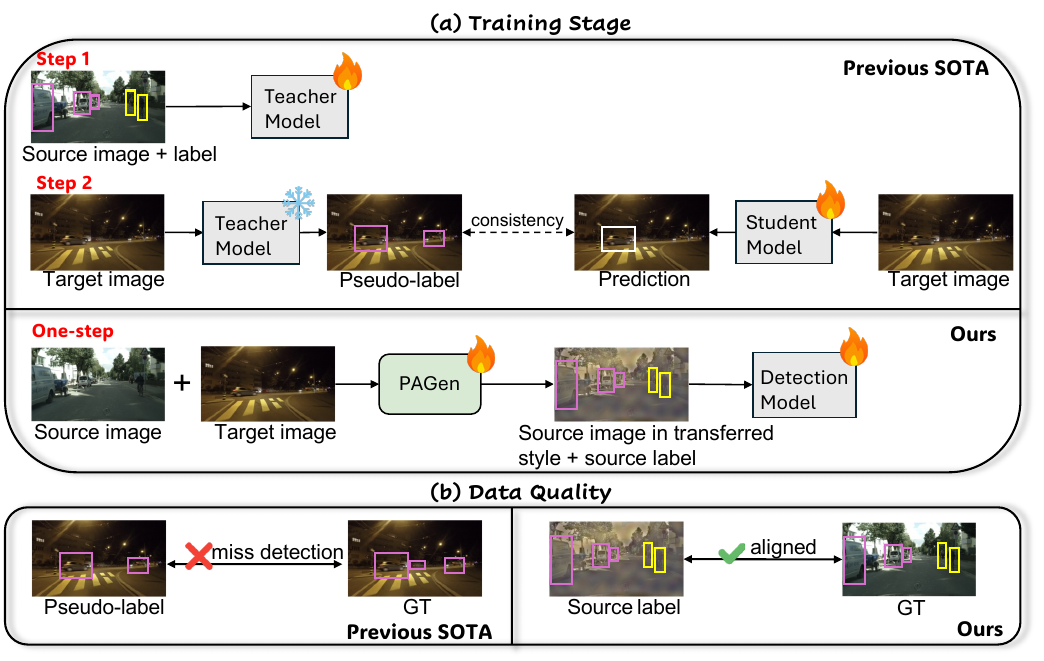}
  \caption{(a) Recent state-of-the-art approaches mainly rely on self-training strategies to generate pseudo labels for target-domain images. Their training procedures are typically complex, often requiring multi-stage optimization and careful balancing among multiple loss terms. In contrast, our method enables a simple end-to-end one-step training pipeline, which greatly facilitates practical deployment.
(b) In terms of target-domain data quality, existing methods heavily rely on pseudo labels for consistency training, which can introduce noise due to missed detections. Our approach, however, ensures accurate target supervision without such noise, thus reducing misleading gradients during adaptation.}
  \label{fig:arch-cmp}
\end{figure}


Object detection is widely applied in real-world scenarios such as surveillance, autonomous driving, and robotics. Despite its success, training robust detectors for every new environment remains challenging, as collecting high-quality bounding box annotations for each scenario is both costly and labor-intensive. Unsupervised domain adaptation (UDA) addresses this challenge by transferring knowledge from a labeled source domain to an unlabeled target domain, thereby reducing the annotation burden while improving generalization across diverse domains.

Pioneering works on domain-adaptive object detection (DAOD) employ adversarial losses~\cite{Chen_2018_CVPR, Saito_2019_CVPR, Chen_2020_CVPR, He_2019_ICCV, Xu_2020_CVPR, Zhu_2019_CVPR, su2020adapting} or graph matching~\cite{Li_Liu_Yao_Yuan_2022, 10204361, 10012542, Li_2022_CVPR, Liu_2022_CVPR, 10376923} for domain alignment. However, these methods often bias the detector toward the source domain, as only source-domain images are labeled during training, limiting their effectiveness. As a solution, recent approaches adopt the self-training strategy to generate pseudo labels for the target domain~\cite{pmlr-v162-chen22b, li2022cross, 10204157, pmlr-v235-feng24d, Deng_2021_CVPR, 10658175, Kennerley_2023_CVPR}, implemented either via a mean-teacher framework~\cite{NIPS2017_68053af2, Deng_2021_CVPR} or using vision foundation models (VFM)~\cite{Lavoie_2025_CVPR, Li_2025_CVPR, NEURIPS2023_0d18ab3b}. Although these methods have progressively improved performance, their implementation procedures are often complicated (e.g., requiring careful balancing of multiple losses in a multi-task setting) or resource-intensive (e.g., multi-stage training including large auxiliary models, or VFM with copyright restrictions), limiting their practicality.

We aim to develop a simple yet effective approach for DAOD. To minimize additional computational overhead and reduce dependency on auxiliary networks as mentioned above, we tackle the problem at the input level. Specifically, we draw inspiration from frequency-domain image decomposition~\cite{gonzalez_digital_2018}: In the frequency domain, the discrete Fourier transform (DFT) represents an image with its phase and amplitude components, where the phase encodes content and structural layout on which the human perception relies more for semantic recognition, and the amplitude reflects the style and noise that are more domain-sensitive~\cite{Yang_2020_CVPR, Chen_2021_ICCV}. Based on this property, re-pairing the phase and amplitude components from different images enables the recomposition of content and style, improving model robustness against amplitude variations in different domains.

To this end, we introduce Phase-guided Amplitude Generation (PAGen), a lightweight (only $120k$ parameters) learnable module that performs input-level style adaptation for the DAOD task. Given labeled source image data and the unlabeled target one, PAGen operates in the frequency domain by decomposing images into content and style components, and learns to transfer target-domain styles to source images while preserving the original content and labels therein. Furthermore, unlike existing approaches that require auxiliary models or multi-stage training pipelines~\cite{Deng_2021_CVPR, li2022cross, 10204157, Lavoie_2025_CVPR, pmlr-v162-chen22b, 10203852}, PAGen can be inserted (and removed in the inference phase) as a simple preprocessing module and trained end-to-end together with the downstream task in a single training stage, significantly simplifying the training process. Our main contributions can be summarized as follows:
\begin{itemize}
    \item We introduce PAGen, a simple, efficient, yet effective UDA method that incorporates only a preprocessing module during training, which is fully discarded at inference to avoid any additional computational overhead. 
    \item By inherently maintaining content consistency between style-adapted and source images, we propose a feature alignment loss to promote robust feature learning and boost performance.
    \item Comprehensive experiments on various DAOD benchmarks demonstrate the effectiveness of our approach, achieving consistently superior performance with improvements of $2.3\%$ on Foggy Cityscapes, $3.1\%$ on BDD100K Night, and an average of $1.7\%$ on ACDC over state-of-the-art complicated methods.
\end{itemize}
\section{Related works}
\label{sec:related_work}

We briefly review existing works on domain-adaptive object detection and domain robust learning in the frequency domain.

\subsection{Domain-adaptive object detection}
Early works on domain-adaptive object detection primarily rely on adversarial learning~\cite{pmlr-v37-ganin15} to align feature distributions at both the image and object levels~\cite{Chen_2018_CVPR, Saito_2019_CVPR, Chen_2020_CVPR, He_2019_ICCV, Xu_2020_CVPR, Zhu_2019_CVPR, su2020adapting}. Another line of research models domain distributions with graph structures and reduces domain discrepancies via graph matching~\cite{Li_Liu_Yao_Yuan_2022, 10204361, 10012542, Li_2022_CVPR, Liu_2022_CVPR, 10376923}. For instance, DA-Faster~\cite{Chen_2018_CVPR} aligns feature distributions at both the image and object levels, while SIGMA~\cite{Li_2022_CVPR} models source and target domains as graphs and performs domain adaptation via graph matching.  However, due to the absence of ground-truth annotations in the target domain, the detector becomes biased toward the source domain and consequently exhibits limited performance on the target domain. To mitigate this issue, self-training methods have been proposed to leverage pseudo labels~\cite{pmlr-v162-chen22b, li2022cross, 10204157, pmlr-v235-feng24d, Deng_2021_CVPR, 10658175, Kennerley_2023_CVPR}, generally following a mean-teacher framework~\cite{NIPS2017_68053af2, Deng_2021_CVPR} with various data augmentations. For example, UMT~\cite{Deng_2021_CVPR} employs CycleGAN~\cite{CycleGAN2017}-generated images to mitigate domain shifts. CMT~\cite{10204157} integrates mean-teacher with contrastive learning~\cite{He_2020_CVPR} to enhance the representative capability of pseudo labels. AT~\cite{li2022cross} combines adversarial domain learning with weak–strong augmentation. MTOR~\cite{8953637} employs mean-teacher to enforce region- and graph-level consistency. More recently, with the rise of VFMs, DT~\cite{Lavoie_2025_CVPR} utilizes DINOv2~\cite{oquab2024dinov} for offline pseudo-label generation and online feature distillation. SEEN-DA~\cite{Li_2025_CVPR} employs semantic entropy to refine cross-domain features using CLIP’s textual encoder~\cite{radford2021learning} and RegionCLIP’s visual encoder~\cite{Zhong_2022_CVPR}. Despite their effectiveness, self-training methods often require multiple training stages and rely on threshold-based~\cite{Deng_2021_CVPR, li2022cross} or complex pseudo-label selection criteria~\cite{Deng_2023_CVPR, pmlr-v162-chen22b}, making them difficult to train and tune in practice.

\subsection{Domain-robust learning in frequency domain} 
Images can be transformed into the frequency domain via various methods. For instance, the complex-valued discrete Fourier transform (DFT)~\cite{gonzalez_digital_2018} maps images into the frequency domain, which can be decomposed into phase and amplitude components, facilitating the style transfer of images by the manipulation of the amplitude spectrum. FDA~\cite{Yang_2020_CVPR} exploits this property and swaps the low-frequency bands of amplitude spectra between source and target images for domain-adaptive semantic segmentation. ARP~\cite{Chen_2021_ICCV} swaps the entire amplitude spectrum to enhance feature robustness. More recently, CISS~\cite{10840277} applies FDA to adapt between normal and adverse weather conditions for segmentation. Despite their efficiency in transferring style for DA, these methods rely on a fixed hyperparameter to determine which frequency bands are exchanged. This lack of adaptability across different domains or image instances prevents the model from fully realizing its performance potential. There also exist real-valued transformations, such as the discrete sine transform (DST)~\cite{106875} and the discrete cosine transform (DCT)~\cite{1672377}, which do not support phase–amplitude decomposition well. In these cases, the spectral signals are typically analyzed manually and partitioned into domain-invariant and domain-variant frequency bands, which are then adapted separately in the latent frequency domain. This strategy is grounded in the observation that low-frequency components tend to be more domain-invariant and generalizable, while high-frequency components are often domain-specific and prone to overfitting~\cite{Lin_2023_CVPR, Wang_2020_CVPR, Chen_2021_ICCV}. For example, FSDR~\cite{Huang_2021_CVPR} and NightAdapter~\cite{Bi_2025_CVPR} leverage DCT and DST, respectively, to enhance domain-invariant features while attenuating domain-variant components for domain-generalizable semantic segmentation. However, because these methods rely on manual spectral analysis and handcrafted frequency partitioning, they are susceptible to human bias and result in suboptimal overall performance.

\section{Proposed method}
\label{sec:methodology}

We start from reviewing the problem setup of UDA for object detection and the mechanism of image decomposition in the frequency domain for self-completeness. We then present the mechanism of our proposed PAGen and detail its integration into the DAOD task.


\subsection{Preliminaries}

\noindent\textbf{UDA for object detection} 
In the setting of UDA for object detection, we have a labeled source dataset $D^s=\{(x^s_i, b^s_i, c^s_i) \sim P(x^s, b^s, c^s)\}^{N_s}_{i=1}$, where $x^s_i$ is the $i$-th of the $N_s$ source images; $b^s_i$ and $c^s_i$ represent the corresponding labels of the bounding boxes and categories, respectively; $s$ indicates the source dataset. We also have access to an unlabeled target data set $D^t=\{x^t_i \sim P(x^t)\}^{N_t}_{i=1}$, where $x^t_i$ is the $i^{\text{th}}$ of the target images $N_t$, and $t$ indicates the target data set. In general, an object detector trained on $D^s$ will have a performance drop when tested on $D^t$, especially if the domain gap between the two datasets is large. Here, we propose PAGen to reduce that gap.

%

\begin{figure}[t]
 \centering
 \includegraphics[width=\linewidth]{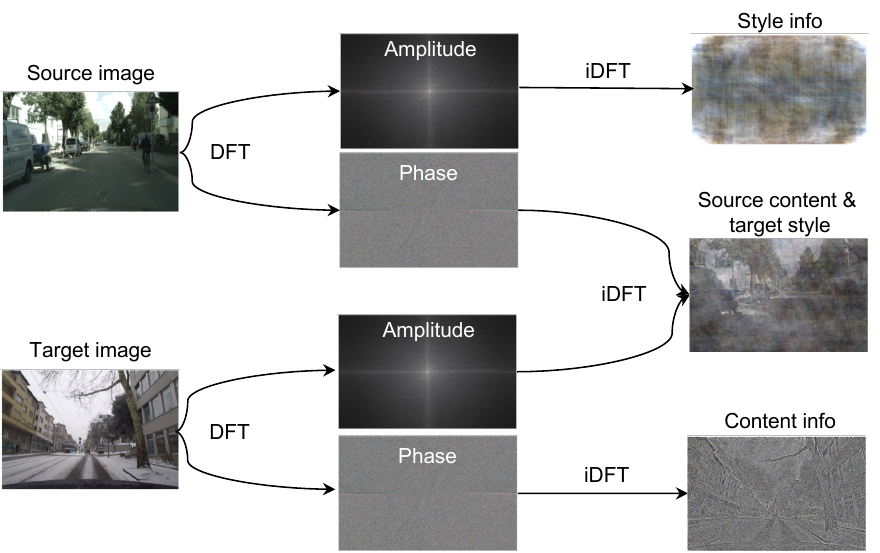}
 \caption{Image decomposition in the frequency domain allows separating content and style components for cross-image recombination.}
 \label{fig:fda}
\end{figure}

\paragraph{Phase-amplitude decomposition of an image} 
An image $I$ of size $H \times W$ can be decomposed into phase and amplitude components after being transformed from the spatial to the frequency domain. Specifically, for each channel of the image, we can convert it from the real spatial domain to the complex frequency domain using the discrete Fourier transform (DFT):
\begin{equation}
\label{eq:1}
\mathcal{F}\left(u,v\right) = \sum_{h, w}^{H,W} I\left(h,w\right)e^{-j2\pi\left(\frac{h}{H}u + \frac{w}{W}v\right)},
\end{equation}
where $j^2=-1$ and $\left(u, v\right)$ denotes the image coordinate. Then we can have the phase spectrum, the argument of $\mathcal{F}$ at each frequency by:
\begin{equation}
\label{eq:2}
\mathcal{F}^{\mathcal{P}}\left(u,v\right)=\mathsf{arctan}\left(\frac{\mathrm{Imag}\left(\mathcal{F}\left(u,v\right)\right)}{\mathrm{Real}\left(\mathcal{F}\left(u,v\right)\right)}\right),
\end{equation}
and amplitude spectrum, the absolute value of $\mathcal{F}$ at each frequency by
\begin{equation}
\label{eq:3}
\mathcal{F}^{\mathcal{A}}\left(u,v\right)=\sqrt{\mathrm{Real}\left(\mathcal{F}\left(u,v\right)\right)^2 + \mathrm{Imag}\left(\mathcal{F}\left(u,v\right)\right)^2},
\end{equation}
where $\mathrm{Real}\left(\cdot\right)$ and $\mathrm{Imag}\left(\cdot\right)$ represent the real and imaginary parts of the complex $\mathcal{F}$, respectively. We use $\mathcal{F}^{-1}$ to represent the discrete inverse Fourier transform ($\mathrm{iDFT}$) that maps the spectrum back to the image domain. Perceptually, $\mathcal{F}^{\mathcal{P}}$ represents the content, while $\mathcal{F}^{\mathcal{A}}$ represents the style of an image, as illustrated in Fig.~\ref{fig:fda}. 


\subsection{Phase-guided Amplitude Generation}



\begin{figure*}[t]
  \centering
  \includegraphics[width=\textwidth,height=3.5cm]{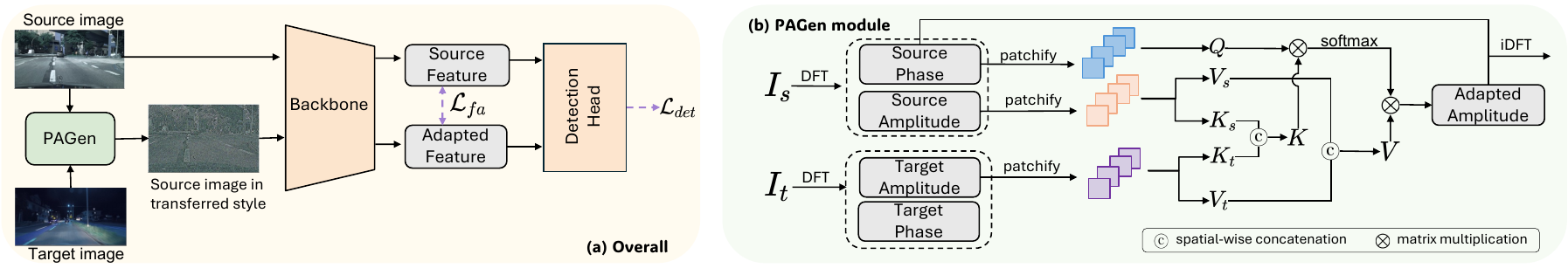}
  \caption{(a) Overview of our approach. The proposed PAGen is employed only during training as a preprocessing component for the detector. Given a source image and its style-altered PAGen output, we feed both into the detector and enforce feature alignment within the backbone’s latent space. This alignment is optimized jointly with the detector’s standard training objectives to learn the full pipeline. During inference time, PAGen is entirely discarded, and target-domain images are directly fed into the detector. (b) Architecture of PAGen. PAGen follows a patch-wise cross-attention paradigm. We transform source image $I_s$ and target image $I_t$ into frequency domain, use the source phase as the query branch, and construct key–value features by concatenating amplitude representations from both domains along the spatial dimension. Cross-attention yields an adapted amplitude, which is combined with the source phase and converted back via iDFT to produce the PAGen-adapted image. }
  
  \label{fig:2}
\end{figure*}

As illustrated in \fref{fig:2}(a), our proposed PAGen operates as a preprocessing module preceding the detection backbone. It is activated only during training, where it jointly processes paired source–target images. During inference, the PAGen module is entirely discarded, and target-domain images are directly fed into the detector. Specifically, We build PAGen upon a patch-based cross-attention framework, where the query is derived from the source phase, while the key and value are obtained from the source and target amplitudes, respectively. In this design, each style-free query patch, representing diverse object or background structures, learns to refine its style through the powerful global attention across both source and target amplitudes, narrowing the domain gap consequently.

As shown in Fig.~\ref{fig:2}(b), starting from two randomly paired input images $I_s$ and $I_t$ from the source and target datasets, respectively, we calculate the phase spectrum of the source image $\mathcal{F}^\mathcal{P}_s$ with Eq.~\ref{eq:2}, and the amplitude spectra $\mathcal{F}^\mathcal{A}_s$ and $\mathcal{F}^\mathcal{A}_t$ of both source and target images with Eq.~\ref{eq:3}, and use them as input to PAGen, as shown in Fig.~\ref{fig:2}(a). In the cross-attention framework, we use $\mathcal{F}^\mathcal{P}_s$ to generate the query \hbox{$Q$}: 
\begin{equation}
\label{eq:4}
Q = \mathrm{DConv}^{3 \times 3}\left(\mathrm{Conv}^{p \times p}\left(\mathcal{F}^\mathcal{P}_s\right)\right),
\end{equation}
in which \hbox{$\mathcal{F}^\mathcal{P}_s$} is first patchified by a convolution operator $\mathrm{Conv}^{p \times p}$ with both kernel size and stride of $p$ (divisible by $H$ and $W$), and simultaneously mapped to hidden dimension $d_h$. Then, a depthwise convolution~\cite{Howard2017MobileNetsEC} with kernel size $3$ is used to project it to the query space. Similarly, we use \hbox{$\mathcal{F}^\mathcal{A}_{s}$} and \hbox{$\mathcal{F}^\mathcal{A}_{t}$} to generate keys \hbox{$K_{s},\ K_{t}$} and values \hbox{$V_{s}, \ V_{t}$} from the source and target amplitude spectrum, respectively:
\begin{equation}
\label{eq:5}
\begin{aligned}
K_{s} &= \mathrm{DConv}^{3 \times 3}\left(\mathrm{Conv}^{p \times p}\left(\mathcal{F}^\mathcal{A}_{s}\right)\right), \\
K_{t} &= \mathrm{DConv}^{3 \times 3}\left(\mathrm{Conv}^{p \times p}\left(\mathcal{F}^\mathcal{A}_{t}\right)\right),
\end{aligned}
\end{equation}
\begin{equation}
\label{eq:6}
\begin{aligned}
V_{s} = \mathrm{DConv}^{3 \times 3}\left(\mathrm{Conv}^{p \times p}\left(\mathcal{F}^\mathcal{A}_{s}\right)\right), \\
V_{t} = \mathrm{DConv}^{3 \times 3}\left(\mathrm{Conv}^{p \times p}\left(\mathcal{F}^\mathcal{A}_{t}\right)\right).
\end{aligned}
\end{equation}
It is noteworthy that the convolution layers in Eqs.~(\ref{eq:4}--\ref{eq:6}) are not shared. We then flatten the keys and values in the spatial dimensions and concatenate $K_s$ and $K_t$, $V_s$ and $V_t$ pairwisely in the flattened spatial dimension to formulate the key \hbox{$K$} and value \hbox{$V$} for the cross-attention framework:
\begin{equation}
\label{eq:7}
K = \mathrm{concat}\left(K_s, K_t\right),
\end{equation}
\begin{equation}
\label{eq:8}
V = \mathrm{concat}\left(V_s, V_t\right).
\end{equation}
We also flatten $Q$ in \eref{eq:4} in the spatial dimensions and generate output amplitude \hbox{$A$} via the computation of the attention map~\cite{Zamir_2022_CVPR}:
\begin{equation}
\label{eq:9}
A = \mathrm{Softmax}\left(\frac{QK^T}{t}\right)V,
\end{equation}
in which $t$ is a learnable temperature to scale the sharpness of the attention distribution. Here, we assume a single attention head for clarity, without loss of generality. We then reshape $A$ back to separate spatial dimensions and generate the final output amplitude \hbox{$A^{\ast}$}:
\begin{equation}
\label{eq:10}
A^{\ast} = \mathrm{Conv}^{1 \times 1}\left(\mathrm{Upsample}\left(A\right)\right),
\end{equation}
in which $\mathrm{Upsample}$ is a bilinear interpolator that extends the size back to $H \times W$, and a point-wise convolution~\cite{Howard2017MobileNetsEC} $\mathrm{Conv}^{1 \times 1}$ is used to transform it back to the input channel $C$. Now we can get a style-transferred image whose content is consistent with the source image, by using $\mathrm{iDFT}$ to map $\mathcal{F}^{\mathcal{P}}_s$ and $A^{\ast}$ back to image space:
\begin{equation}
\label{eq:11}
I_a = \mathcal{F}^{-1}\left(\mathcal{F}^{\mathcal{P}}_s, A^{\ast}\right),
\end{equation}
in which $I_a$ is the adapted image from the source domain to the target. It is used together with the source image $I_{s}$ for model training to enhance feature robustness between source and target domains.

\subsection{Feature alignment} 
To strengthen the consistency of learned features across the source and target domains, we additionally introduce a feature alignment loss, as illustrated in Fig.~\ref{fig:2}(a). This loss encourages alignment of the multi-stage backbone feature maps, facilitating the learning of domain-invariant representations:
\begin{equation}
\label{eq:loss-fa}
\mathcal{L}_{fa} = \frac{1}{N}\sum_{i=1}^{N} \mathrm{MSE}\left(F^s_i, F^a_i\right),
\end{equation}
where the mean squared error $\mathrm{MSE}$ is computed between the source and adapted feature maps of the $i^{\text{th}}$ stage of the backbone $F^s_i$, $F^a_i$, and averaged among all the $N$ stages. In contrast to previous methods~\cite{Chen_2018_CVPR, Saito_2019_CVPR} that align domains through adversarial learning at the distribution level, our approach directly adapts and aligns corresponding feature samples, achieving a more fine-grained and effective feature alignment.

\begin{table*}[t]
  \caption{Domain adaptive object detection from Cityscapes to Foggy Cityscapes. The best performances are in {\bf bold}, and the second-best are \underline{underlined}. The corresponding technical properties of each method are listed as: Alignment (adversarial domain alignment and graph matching), Self-training (online pseudo-label generation or self-distillation), and VFM-based (offline pseudo-label generation, feature distillation, or fine-tuning).}
  \label{tab:cs2foggy}
  \centering
  \begin{adjustbox}{max width=\linewidth}
  \begin{tabular}{lcccccccccccc}
    \toprule

    \multirow{2}{*}{Method} & \multicolumn{3}{c}{Technical properties} & \multirow{2}{*}{Person} & \multirow{2}{*}{Rider} & \multirow{2}{*}{Car} & \multirow{2}{*}{Truck} & \multirow{2}{*}{Bus} & \multirow{2}{*}{Train} & \multirow{2}{*}{Motor} & \multirow{2}{*}{Bicycle} & \multirow{2}{*}{mAP} \\
    \cmidrule(lr){2-4} 
  & Alignment & Self-training & VFM-based \\

    \midrule
    DA-Faster~\cite{Chen_2018_CVPR} & \checkmark & & & 29.2 & 40.4 & 43.4 & 19.7 & 38.3 & 28.5 & 23.7 & 32.7 & 32.0 \\
    DICN~\cite{9935311} & \checkmark & & & 47.3 & 57.4 & 64.0 & 22.7 & 45.6 & 29.6 & 38.6 & 47.4 & 44.1 \\
    NLTE~\cite{Liu_2022_CVPR} & \checkmark & & & 43.1 & 50.7 & 58.7 & 33.6 & 56.7 & 42.7 & 33.7 & 43.3 & 45.4 \\
    \midrule
    PT~\cite{pmlr-v162-chen22b} & & \checkmark & & 43.2 & 52.4 & 63.4 & 33.4 & 56.6 & 37.8 & 41.3 & 48.7 & 47.1 \\
    CMT~\cite{10204157} & & \checkmark & & 47.0 & 55.7 & 64.5 & 39.4 & 63.2 & 51.9 & 40.3 & 53.1 & 51.9 \\
    \midrule
    MIC~\cite{10203852} & \checkmark & \checkmark & & 50.9 & 55.3 & 67.0 & 33.9 & 52.4 & 33.7 & 40.6 & 47.5 & 47.6 \\
    AT~\cite{li2022cross} & \checkmark & \checkmark & & 45.5 & 55.1 & 64.2 & 35.0 & 56.3 & \underline{54.3} & 38.5 & 51.9 & 50.9 \\
    DSD-DA~\cite{pmlr-v235-feng24d} & \checkmark & \checkmark & & 49.0 & 59.6 & 65.3 & 35.7 & 61.0 & 46.5 & 43.9 & 57.3 & 52.3 \\
    CAT~\cite{10658175} & \checkmark & \checkmark & & 44.6 & 57.1 & 63.7 & 40.8 & \underline{66.0} & 49.7 & 44.9 & 53.0 & 52.5 \\
    NSA-UDA~\cite{10376923} & \checkmark & \checkmark & & 50.3 & 60.1 & 67.7 & 37.4 & 57.4 & 46.9 & 47.3 & 54.3 & 52.7 \\
    REACT~\cite{10552660} & \checkmark & \checkmark & & 51.4 & 57.9 & 67.4 & 37.7 & 58.4 & 52.8 & 44.6 & 54.6 & 53.1 \\
    \midrule
    DT~\cite{Lavoie_2025_CVPR} & & \checkmark & \checkmark & 48.5 & 60.0 & 65.4 & \textbf{47.2} & \textbf{66.5} & 52.9 & 46.2 & 56.7 & 55.4 \\
    \midrule
    DA-Pro~\cite{NEURIPS2023_0d18ab3b} & \checkmark & \checkmark & \checkmark & 55.4 & 62.9 & 70.9 & 40.3 & 63.4 & 54.0 & 42.3 & 58.0 & 55.9 \\
    SEEN-DA~\cite{Li_2025_CVPR} & \checkmark & \checkmark & \checkmark & 58.5 & 64.5 & 71.7 & 42.0 & 61.2 & \textbf{54.8} & 47.1 & 59.9 & \underline{57.5} \\
    \midrule
    PAGen (Ours) & & & & \textbf{63.7} & \textbf{65.5} & \textbf{74.3} & \underline{46.1} & 63.9 & 49.9 & \textbf{50.8} & \textbf{64.3} & \textbf{59.8} \\
    \bottomrule
  \end{tabular}
  \end{adjustbox}
\end{table*}

\subsection{PAGen in domain-adaptive object detection}
We now describe how to incorporate PAGen into DAOD. PAGen is applied only during training as a preprocessing module and is discarded at inference, ensuring zero additional computational overhead.

\paragraph{PAGen as a pre-processing module} 
As illustrated in Fig.~\ref{fig:2}(b), given one source-domain image and a target-domain counterpart, PAGen generates the adapted image, which is fed into the backbone alongside the original source image. Since the adapted image preserves the content of the source image, the source ground-truth annotations can be directly used to supervise the adapted image. Accordingly, the object detection loss for both the source and adapted images can be expressed as (without loss of generality, we here adopt Faster R-CNN~\cite{NIPS2015_14bfa6bb} for illustration):
\begin{equation}
\label{eq:loss-det}
\mathcal{L}_{det} = \mathcal{L}^{rpn}_s + \mathcal{L}^{roi}_s + \mathcal{L}^{rpn}_a + \mathcal{L}^{roi}_a,
\end{equation}
where $\mathcal{L}^{rpn}_{s}$, $\mathcal{L}^{rpn}_{a}$, $\mathcal{L}^{roi}_{s}$, and $\mathcal{L}^{roi}_{a}$ denote the region proposal loss from the region proposal network, and box regression and classification loss from the region-of-interest layer, on the source and adapted images, respectively. We find that these losses have similar ranges and do not need to be weighted against each other. With the style adaptation at the input level, the model can learn robust and transferable features for testing in the target domain.

\paragraph{Overall optimization objective}
Combining the object detection loss of Eq.~\ref{eq:loss-det} and the feature alignment loss of Eq.~\ref{eq:loss-fa} leads to our overall optimization objective in the form of
\begin{equation}
\label{eq:loss-both}
\mathcal{L}_{PAGen} = \mathcal{L}_{det} + \lambda\mathcal{L}_{fa},
\end{equation}
where $\lambda$ helps to balance the two losses.



\section{Experiments}
We extensively compare our proposal with state-of-the-art methods on various datasets and present comprehensive ablation studies.

\subsection{Datasets}
We employ five popularly used dataset for comparison:
\begin{itemize}
    \item \textbf{ACDC~\cite{Sakaridis_2021_ICCV}} is a real-world autonomous driving dataset covering four adverse weather conditions: fog, night, rain, and snow. Its object detection annotations are consistent with Cityscapes~\cite{7780719}. Each weather condition includes $400$ training images and $100$ validation images (except for night, which has $106$ validation images). Following prior work~\cite{Lavoie_2025_CVPR}, we train and evaluate our models separately on each of the four weather splits.
    \item \textbf{Cityscapes~\cite{7780719}} is a real-world autonomous driving dataset comprising $2,975$ training images and $500$ validation images for $8$ classes.
    \item \textbf{Foggy Cityscapes~\cite{Sakaridis2017SemanticFS}} is a synthetic dataset derived from Cityscapes that simulates fog at three intensity levels ($0.005$, $0.01$, $0.02$). Its annotations are inherited from Cityscapes. The dataset contains $8,925$ images for training and $1,500$ for validation. Following prior work~\cite{Lavoie_2025_CVPR, Li_2025_CVPR}, we use the full dataset in our experiments.
    \item \textbf{BDD100K~\cite{9156329}} is a real-world autonomous driving dataset with annotations for $10$ classes. Following prior works~\cite{Kennerley_2023_CVPR}, we use the day and night training splits, which contain $36,728$ and $27,971$ images, respectively, and the night validation split, which contains $3,929$ images.
    \item \textbf{Sim10k~\cite{JohnsonRoberson2016DrivingIT}} is a synthetic dataset generated from the GTA V game\footnote{Rockstar Games, Grand Theft Auto V, Rockstar North, 2013.}. It contains $10,000$ images with the class \textit{car} for object detection labels.
    
\end{itemize}

\subsection{Implementation details}

We implement our method based on MMDetection~\cite{mmdetection}. We adopt ResNet50~\cite{He_2016_CVPR} pre-trained on COCO~\cite{Lin2014MicrosoftCC} as the backbone and use Faster R-CNN~\cite{NIPS2015_14bfa6bb} as the detection framework. In general, we follow the default training configurations in MMDetection, except that we set the batch size to $2$ to include one source image and one target image per iteration. All experiments are conducted on an H100 GPU.

For the hyper-parameters in PAGen, we set the patch size to $16$ and resize input images such that their spatial dimensions are divisible by this patch size. We set the number of heads to $4$, the number of hidden dimensions to $32$, and the loss weight $\lambda$ in Eq.~\ref{eq:loss-both} to $1$.


\subsection{Results and comparisons}
\label{subsec:res_and_comp}
We employ the mean average precision (mAP) with a threshold of $0.5$ as the evaluation metric. All experimental setups are listed below:

\begin{itemize}
    \item \textbf{Cityscapes $\to$ Foggy Cityscapes} measures adaptation from real-world clear weather to synthetic foggy conditions. As shown in Table~\ref{tab:cs2foggy}, PAGen surpasses all compared methods in terms of mAP, improving the state-of-the-art from $57.5\%$ to $59.8\%$ (+$2.3\%$). In particular, our method outperforms $5$ of the $8$ categories (person, rider, car, motor, and bicycle), with gains ranging from $1\%$ to $5.2\%$.
    \item \textbf{Sim10k $\to$ Cityscapes} assesses adaptation from synthetic images generated by the GTA V game engine to real-world driving scenes. As shown in Table~\ref{tab:sim2cs}, PAGen improves the state-of-the-art performance to $69.9\%$ mAP, achieving a gain of $3.1\%$.
    \item \textbf{BDD100K Day $\to$ Night} evaluates adaptation from real-world daytime to nighttime driving scenes. As shown in Table~\ref{tab:d2n}, PAGen achieves the best performance of $49.5\%$, improving upon the current state-of-the-art method by $3.1\%$. Following prior work, we exclude the \textit{train} class from evaluation due to its limited annotations. Notably, our method outperforms $7$ out of the $9$ evaluated classes, with performance gains ranging from $2.2\%$ to $8.9\%$.
    \item \textbf{Cityscapes $\to$ ACDC} evaluates adaptation from real-world normal weather to adverse weather conditions. As shown in Table~\ref{tab:cs2acdc}, PAGen achieves performance improvements in $3$ out of the $4$ adverse weather settings, with gains ranging from $1.2\%$ to $4.6\%$. On average, PAGen improves performance by $1.7\%$ across all four conditions.
\end{itemize}

\begin{table}[t]
  \caption{Domain adaptive object detection from Sim10k to Cityscapes.}
  \label{tab:sim2cs}
  \centering
  \begin{adjustbox}{max width=\linewidth}
  \begin{tabular}{lcccc}
    \toprule
    \multirow{2}{*}{Method} & \multicolumn{3}{c}{Technical properties} & \multirow{2}{*}{mAP} \\
\cmidrule(lr){2-4} 
  & Alignment & Self-training & VFM-based \\
  
    \midrule
    LRA~\cite{10154573} & \checkmark & & & 55.7 \\
    DA-Faster~\cite{Chen_2018_CVPR} & \checkmark & & & 38.2 \\ 
    SCAN~\cite{Li_Liu_Yao_Yuan_2022} & \checkmark & & & 52.6 \\
    CIGAR~\cite{10204361} & \checkmark & & & 58.5 \\
    OADA~\cite{10.1007/978-3-031-19827-4_40} & \checkmark & & & 59.2 \\
    
    \midrule
    DSD-DA~\cite{pmlr-v235-feng24d} & \checkmark & \checkmark & & 52.5 \\
    NSA-UDA~\cite{10376923} & \checkmark & \checkmark & & 56.3 \\
    SIGMA++~\cite{10012542} & \checkmark & \checkmark & & 57.7 \\
    REACT~\cite{10552660} & \checkmark & \checkmark & & 58.6 \\
    SOCCER~\cite{cui2024stochastic} & \checkmark & \checkmark & & 63.8 \\
    
    \midrule
    DA-Pro~\cite{NEURIPS2023_0d18ab3b} & \checkmark & \checkmark & \checkmark & 62.9 \\
    SEEN-DA~\cite{Li_2025_CVPR} & \checkmark & \checkmark& \checkmark & \underline{66.8} \\
    \midrule
    PAGen (Ours) & & & & \textbf{69.9} \\
    \bottomrule
  \end{tabular}
  \end{adjustbox}
\end{table}

\begin{table*}[t]
  \caption{Results of domain adaptive object detection from BDD100K Day to Night.}
  \label{tab:d2n}
  \centering
  \begin{adjustbox}{max width=\linewidth}
  \begin{tabular}{lcccccccccccc}
    \toprule

    \multirow{2}{*}{Method} & \multicolumn{2}{c}{Technical properties} & \multirow{2}{*}{Pedestrian} & \multirow{2}{*}{Rider} & \multirow{2}{*}{Car} & \multirow{2}{*}{Truck} & \multirow{2}{*}{Bus} & \multirow{2}{*}{\makecell{Motor \\ cycle}} & \multirow{2}{*}{Bicycle} & \multirow{2}{*}{\makecell{Traffic \\ Light}} & \multirow{2}{*}{\makecell{Traffic \\ Sign}} & \multirow{2}{*}{mAP} \\
    \cmidrule(lr){2-3} 
  & Alignment & Self-training \\
    \midrule
    DA-Faster~\cite{Chen_2018_CVPR} & \checkmark & & 50.4 & 30.3 & 66.3 & 46.8 & 48.3 & 32.6 & 41.4 & 41.0 & 56.2 & 41.3 \\
    \midrule
    TDD~\cite{He_2022_CVPR} & & \checkmark & 43.1 & 20.7 & 68.4 & 33.3 & 35.6 & 16.5 & 25.9 & 43.1 & 59.5 & 34.6 \\
    UMT~\cite{Deng_2021_CVPR} & & \checkmark & 46.5 & 26.1 & 46.8 & 44.0 & 46.3 & 28.2 & 40.2 & 31.6 & 52.7 & 36.2 \\
    2PCNet~\cite{Kennerley_2023_CVPR} & & \checkmark & \underline{54.4} & \underline{30.8} & \underline{73.1} & \textbf{53.8} & \textbf{55.2} & \underline{37.5} & \underline{44.5} & \underline{49.4} & \underline{65.2} & \underline{46.4} \\
    \midrule
    AT~\cite{li2022cross} & \checkmark & \checkmark & 42.3 & 30.4 & 60.8 & 48.9 & \underline{52.1} & 34.5 & 42.7 & 29.1 & 43.9 & 38.5 \\
    \midrule
    PAGen (Ours) & & & \textbf{61.0} & \textbf{39.7} & \textbf{75.3} & \underline{49.7} & 48.9 & \textbf{45.3} & \textbf{48.4} & \textbf{57.0} & \textbf{70.1} & \textbf{49.5} \\
    \bottomrule
  \end{tabular}
  \end{adjustbox}
\end{table*}

\begin{table}[t]
\caption{Domain adaptive object detection from Cityscapes to ACDC on the four weather splits.}
\label{tab:cs2acdc}
\centering
\begin{adjustbox}{max width=\linewidth}
  \begin{tabular}{lccccccc}
    \toprule
    \multirow{2}{*}{Method} & \multicolumn{3}{c}{Technical properties} & \multirow{2}{*}{Fog} & \multirow{2}{*}{Night} & \multirow{2}{*}{Rain} & \multirow{2}{*}{Snow} \\
    \cmidrule(lr){2-4} 
    & Alignment & Self-training & VFM-based \\
  
    \midrule
    AT~\cite{li2022cross} & \checkmark & \checkmark & & 62.2 & 29.5 & 37.7 & 55.2 \\
    DT~\cite{Lavoie_2025_CVPR} & & \checkmark & \checkmark & \textbf{68.6} & \underline{36.4} & \underline{39.0} & \underline{56.8} \\
    \midrule
    \makecell{PAGen\\(Ours)} & & & & \underline{67.8} & \textbf{37.6} & \textbf{43.6} & \textbf{58.6} \\
    \bottomrule
  \end{tabular}
\end{adjustbox}
\end{table}

\subsection{Ablation Studies}
To validate the effectiveness of the proposed method, we conduct ablation studies on the challenging Cityscapes $\rightarrow$ ACDC benchmark, which involves adaptation from clear-weather images to images under various adverse conditions in the real-world scenario.

\begin{table}[t]
  \centering
  \caption{Ablation study \wrt model architectures by domain adaption from Cityscapes to ACDC.}
  \label{tab:ab-ad}
  \begin{adjustbox}{max width=\linewidth}
  \begin{tabular}{ccccccc}
  \toprule
  \multirow{2}{*}{Method} & \multicolumn{2}{c}{Settings} & \multirow{2}{*}{Fog} & \multirow{2}{*}{Night} & \multirow{2}{*}{Rain} & \multirow{2}{*}{Snow} \\ 
  \cmidrule(lr){2-3} 
  & Source amp. & Target amp. \\
  \midrule
  \makecell{FDA~\cite{Yang_2020_CVPR}\\($\beta=0.01$)} & \checkmark & \checkmark & \underline{64.8} & \underline{36.0} & 38.9 & 56.8 \\
  \makecell{FDA~\cite{Yang_2020_CVPR}\\($\beta=0.05$)} & \checkmark & \checkmark & 62.7 & 34.8 & 41.7 & 54.0 \\
  \makecell{FDA~\cite{Yang_2020_CVPR}\\($\beta=0.09$)} & \checkmark & \checkmark & 63.0 & 34.2 & \underline{43.3} & 53.7 \\
  \makecell{PAGen\\(Ours)} &  & \checkmark & 63.3 & 35.1 & 41.8 & \underline{57.1}  \\
  \makecell{PAGen\\(Ours)} & \checkmark & \checkmark & \textbf{67.8} & \textbf{37.6} & \textbf{43.6} & \textbf{58.6}  \\
  \bottomrule
  \end{tabular}
  \end{adjustbox}
\end{table}

\paragraph{Learning-based vs. Human-Tuned Amplitude Generation} 

Starting from the original FDA~\cite{Yang_2020_CVPR}, we progressively evaluate the contributions of the proposed PAGen architecture. Following FDA, we substitute the source amplitude with the target amplitude using $\beta \in \{0.01, 0.05, 0.09\}$. For the design of the cross-attention (CA) module, we first test using only the target amplitude, and then using both source and target amplitudes to generate the key-value pairs. As shown in Table~\ref{tab:ab-ad}, vanilla FDA exhibits variable performance as $\beta$ increases, ranging from a $-3.1\%$ drop (\textit{Snow}) to a $+4.4\%$ gain (\textit{Rain}). This indicates that tuning $\beta$ is necessary to optimize spectral manipulation under different real-world adverse conditions.

\begin{figure*}[t]
  \centering
  \includegraphics[width=\textwidth]{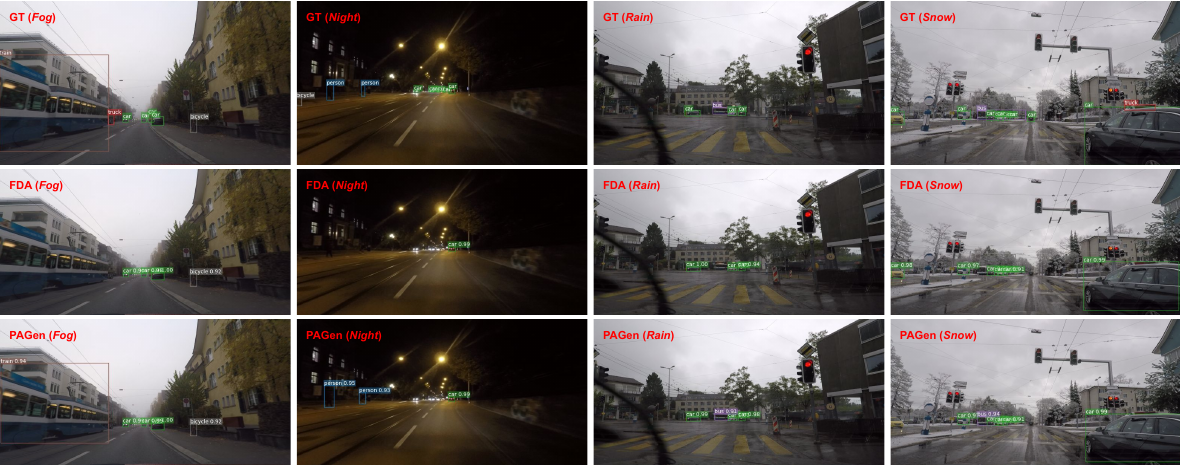}
  \caption{Examples of detection results on each weather split of the ACDC dataset.}
  \label{fig:pred_res}
\end{figure*}

On the other hand, our proposed PAGen outperforms FDA by leveraging the global patch-wise reference of the cross-attention module to learn and generate spectra for domain adaptation. As shown in Table~\ref{tab:ab-ad}, using only the target amplitude in PAGen yields performance comparable to FDA, indicating that incorporating the source amplitude is essential for producing an optimal spectrum, which is consistent with the original motivation of FDA. When introducing the source amplitude, applying cross-attention PAGen yields the best performance and consistently surpasses FDA across all four splits, from $0.3\%$ on \textit{Rain} to $3.0\%$ on \textit{Fog}. These results demonstrate the superiority of leveraging global reference information from both source and target amplitudes simultaneously when predicting the attention map. A qualitative comparison between FDA ($\beta=0.01$) and PAGen on the Cityscapes to ACDC setup across all four adverse conditions is presented in \fref{fig:pred_res}.

\begin{table}[t]
  \caption{Ablation study \wrt different $\lambda$ values for the feature alignment loss by domain adaption from Cityscapes to ACDC.}
  \label{tab:ab-fa}
  \centering
\begin{adjustbox}{max width=\linewidth}
  \begin{tabular}{cccccc}
    \toprule
     & $\lambda=0$ & $\lambda=0.1$ & $\lambda=0.5$ & $\lambda=1$ & $\lambda=5$ \\
    \midrule
    Fog  & 65.3 & \underline{67.2} & 65.8 & \textbf{67.8} & 65.3 \\
    Night & 35.2 & 35.5 & \textbf{37.7} & \underline{37.6} & 35.8 \\
    Rain  & 41.9 & \underline{42.2} & 41.8 & \textbf{43.6} & 40.2 \\
    Snow  & 56.0 & 56.5 & 57.8 & \textbf{58.6} & \underline{58.3} \\
    \bottomrule
  \end{tabular}
\end{adjustbox}
\end{table}

\paragraph{$\lambda$ in the Feature Alignment loss} 
We analyze the effect of varying $\lambda$ values in Eq.~\ref{eq:loss-both}. As presented in Table~\ref{tab:ab-fa}, setting $\lambda=1$ yields the highest performance, with an average improvement of $2.3\%$. Furthermore, all tested $\lambda$ values consistently improve performance over the no-alignment case, emphasizing the crucial role of feature alignment in PAGen.

\begin{table}[t]
  \caption{Ablation studies \wrt the number of attention heads by domain adaption from Cityscapes to ACDC.}
  \label{tab:ab-ah}
  \centering
  \begin{adjustbox}{max width=\linewidth}
  \begin{tabular}{ccccc}
    \toprule
    \makecell{$\#$ Attention heads} & Fog & Night & Rain & Snow \\
    \midrule
    8 & 66.2 & 34.4 & 42.0 & \textbf{59.0} \\
    4 & \textbf{67.8} & \textbf{37.6} & \textbf{43.6} & \underline{58.6} \\
    2 & \underline{67.3} & \underline{37.1} & \underline{43.0} & 58.0 \\
    \bottomrule
  \end{tabular}
  \end{adjustbox}
\end{table}

\paragraph{Number of attention heads} 
We investigate the impact of the number of attention heads on the performance of PAGen. As shown in Table~\ref{tab:ab-ah}, the overall performance generally improves and saturates as the number of attention heads increases from $2$ to $4$, with the exception of the \textit{Snow} condition. While \textit{Snow} sees an additional $0.4\%$ gain when the number of heads increases from $4$ to $8$, the performance on the other three splits decreases, ranging from $1.6\%$ (\textit{Fog} and \textit{Rain}) to $3.2\%$ (\textit{Night}). 
Although varying the number of attention heads triggers slight performance fluctuations, all configurations can achieve comparable or superior performance than the peer methods presented in \Tref{tab:cs2acdc}, demonstrating the effectiveness of our approach.

\begin{table}[t]
  \caption{Ablation studies of domain adaptive object detection from Cityscapes to ACDC on the patch sizes.}
  \label{tab:ab-ps}
  \centering
  \begin{adjustbox}{max width=\linewidth}
  \begin{tabular}{ccccc}
    \toprule
    Patch size & Fog & Night & Rain & Snow \\
    \midrule
    64 & \underline{66.6} & \underline{37.0} & \underline{42.9} & \underline{57.5} \\
    32 & 64.0 & 36.3 & 41.9 & 56.5 \\
    16 & \textbf{67.8} & \textbf{37.6} & \textbf{43.6} & \textbf{58.6}\\
    \bottomrule
  \end{tabular}
  \end{adjustbox}
\end{table}

\paragraph{Influence of patch size} 
We examine the effect of patch size on PAGen’s performance. As shown in Table~\ref{tab:ab-ps}, performance initially decreases when the patch size is reduced from $64$ to $32$, but then improves when the patch size is further reduced to $16$. This result indicates a non-monotonic relationship between model performance and patch size granularity, which we attribute to the trade-off between learning global and local information during training.

\begin{table}[t]
  \caption{Ablation studies of domain adaptive object detection from Cityscapes to ACDC on the number of hidden dimensions.}
  \label{tab:ab-hd}
  \centering
  \begin{adjustbox}{max width=\linewidth}
  \begin{tabular}{ccccc}
    \toprule
    $\#$ Hidden dimensions & Fog & Night & Rain & Snow \\
    \midrule
    64 & \underline{67.3} & 33.2 & 42.4 & \underline{58.1} \\
    32 & \textbf{67.8} & \textbf{37.6} & \textbf{43.6} & \textbf{58.6}\\
    16 & 65.5 & \underline{34.2} & \underline{42.7} & 57.8 \\
    \bottomrule
  \end{tabular}
  \end{adjustbox}
\end{table}

\paragraph{Number of hidden dimensions}
We investigate the impact of varying the number of hidden dimensions on PAGen’s performance. As shown in Table~\ref{tab:ab-hd}, all four splits achieve the highest performance when the number of hidden dimensions is set to $32$. Reducing the hidden dimensions to $16$ or increasing them to $64$ leads to performance drops ranging from $0.5\%$ (\textit{Fog} and \textit{Snow}) to $4.4\%$ (\textit{Night}), likely due to underfitting or overfitting.

\begin{figure}[t]
  \centering
  \begin{subfigure}[t]{0.49\linewidth}
    \centering
    \includegraphics[width=\linewidth]{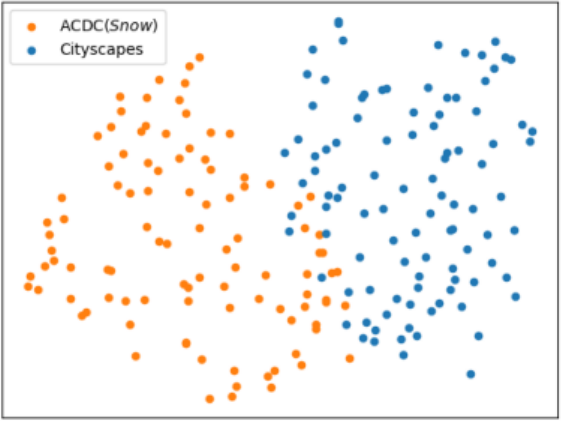}
    \caption{FDA~\cite{Yang_2020_CVPR}}
    \label{fig:first}
  \end{subfigure}
  \hfill
  \begin{subfigure}[t]{0.49\linewidth}
    \centering
    \includegraphics[width=\linewidth]{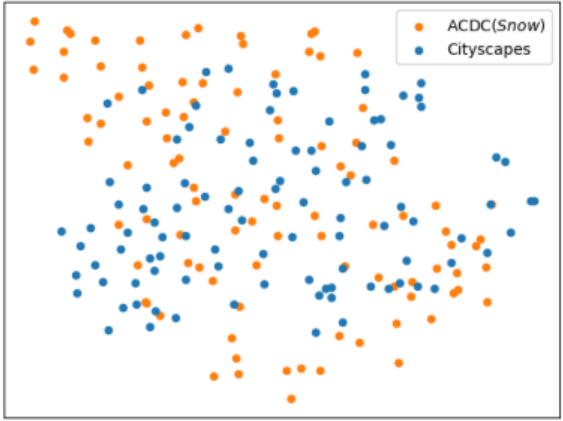}
    \caption{PAGen (ours)}
    \label{fig:second}
  \end{subfigure}
  \caption{t-SNE visualization of embedding features from ACDC (\textit{Snow}) and Cityscapes, illustraing their domain discrepancy and the representational structure learned by the model.}
  \label{fig:t-sne}
\end{figure}

\paragraph{Domain alignment visualization}
We take Cityscapes and ACDC (\textit{Snow}) as an example to illustrate the domain alignment effect. As shown in Fig.~\ref{fig:t-sne}, although also working in the frequency domain, FDA produces feature embeddings of the two datasets that are clearly separated. In contrast, those extracted by PAGen are almost completely overlapped, indicating a stronger robustness against domain difference. This observation exhibits the effectiveness of learning-based design in transferring target-domain styles to source images, leading to the superior performance of PAGen.


\section{Conclusion}
\label{sec:conclusion}
We introduce PAGen, a simple, efficient, yet effective learnable preprocessing module that facilitates unsupervised domain adaptive object detection. Operating in the frequency space, PAGen preserves content-related components from the labeled source domain while adapting style-related components across source and target domains, thereby enhancing the robustness of detectors under domain shifts. Evaluations on multiple DAOD benchmarks demonstrate the superior performance of PAGen over existing approaches.

The main limitation of PAGen lies in addressing content-triggered domain gaps. This is primarily because PAGen relies solely on the phase information from the source domain and does not incorporate phase cues from the target domain. As future work, we plan to investigate phase information fusion to better adapt content-level discrepancies across domains.




{
    \small
    \bibliographystyle{ieeenat_fullname}
    \bibliography{main}

\begin{thebibliography}{60}
\providecommand{\natexlab}[1]{#1}
\providecommand{\url}[1]{\texttt{#1}}
\expandafter\ifx\csname urlstyle\endcsname\relax
  \providecommand{\doi}[1]{doi: #1}\else
  \providecommand{\doi}{doi: \begingroup \urlstyle{rm}\Url}\fi

\bibitem[Ahmed et~al.(1974)Ahmed, Natarajan, and Rao]{1672377}
N. Ahmed, T. Natarajan, and K.R. Rao.
\newblock Discrete cosine transform.
\newblock \emph{IEEE Transactions on Computers}, C-23\penalty0 (1):\penalty0 90--93, 1974.

\bibitem[Bi et~al.(2025)Bi, Yi, Huang, Zheng, Zhan, Huang, Li, Wu, and Zheng]{Bi_2025_CVPR}
Qi Bi, Jingjun Yi, Huimin Huang, Hao Zheng, Haolan Zhan, Yawen Huang, Yuexiang Li, Xian Wu, and Yefeng Zheng.
\newblock Nightadapter: Learning a frequency adapter for generalizable night-time scene segmentation.
\newblock In \emph{Proceedings of the IEEE/CVF Conference on Computer Vision and Pattern Recognition (CVPR)}, pages 23838--23849, 2025.

\bibitem[Cai et~al.(2019)Cai, Pan, Ngo, Tian, Duan, and Yao]{8953637}
Qi Cai, Yingwei Pan, Chong-Wah Ngo, Xinmei Tian, Lingyu Duan, and Ting Yao.
\newblock Exploring object relation in mean teacher for cross-domain detection.
\newblock In \emph{2019 IEEE/CVF Conference on Computer Vision and Pattern Recognition (CVPR)}, pages 11449--11458, 2019.

\bibitem[Cao et~al.(2023)Cao, Joshi, Gui, and Wang]{10204157}
Shengcao Cao, Dhiraj Joshi, Liang-Yan Gui, and Yu-Xiong Wang.
\newblock Contrastive mean teacher for domain adaptive object detectors.
\newblock In \emph{2023 IEEE/CVF Conference on Computer Vision and Pattern Recognition (CVPR)}, pages 23839--23848, 2023.

\bibitem[Chen et~al.(2020)Chen, Zheng, Ding, Huang, and Dou]{Chen_2020_CVPR}
Chaoqi Chen, Zebiao Zheng, Xinghao Ding, Yue Huang, and Qi Dou.
\newblock Harmonizing transferability and discriminability for adapting object detectors.
\newblock In \emph{Proceedings of the IEEE/CVF Conference on Computer Vision and Pattern Recognition (CVPR)}, 2020.

\bibitem[Chen et~al.(2021)Chen, Peng, Ma, Li, Du, and Tian]{Chen_2021_ICCV}
Guangyao Chen, Peixi Peng, Li Ma, Jia Li, Lin Du, and Yonghong Tian.
\newblock Amplitude-phase recombination: Rethinking robustness of convolutional neural networks in frequency domain.
\newblock In \emph{Proceedings of the IEEE/CVF International Conference on Computer Vision (ICCV)}, pages 458--467, 2021.

\bibitem[Chen et~al.(2019)Chen, Wang, Pang, Cao, Xiong, Li, Sun, Feng, Liu, Xu, Zhang, Cheng, Zhu, Cheng, Zhao, Li, Lu, Zhu, Wu, Dai, Wang, Shi, Ouyang, Loy, and Lin]{mmdetection}
Kai Chen, Jiaqi Wang, Jiangmiao Pang, Yuhang Cao, Yu Xiong, Xiaoxiao Li, Shuyang Sun, Wansen Feng, Ziwei Liu, Jiarui Xu, Zheng Zhang, Dazhi Cheng, Chenchen Zhu, Tianheng Cheng, Qijie Zhao, Buyu Li, Xin Lu, Rui Zhu, Yue Wu, Jifeng Dai, Jingdong Wang, Jianping Shi, Wanli Ouyang, Chen~Change Loy, and Dahua Lin.
\newblock {MMDetection}: Open mmlab detection toolbox and benchmark.
\newblock \emph{arXiv preprint arXiv:1906.07155}, 2019.

\bibitem[Chen et~al.(2022)Chen, Chen, Yang, Song, Wang, Zhang, Yan, Qi, Zhuang, Xie, and Pu]{pmlr-v162-chen22b}
Meilin Chen, Weijie Chen, Shicai Yang, Jie Song, Xinchao Wang, Lei Zhang, Yunfeng Yan, Donglian Qi, Yueting Zhuang, Di Xie, and Shiliang Pu.
\newblock Learning domain adaptive object detection with probabilistic teacher.
\newblock In \emph{Proceedings of the 39th International Conference on Machine Learning}, pages 3040--3055. PMLR, 2022.

\bibitem[Chen et~al.(2018)Chen, Li, Sakaridis, Dai, and Van~Gool]{Chen_2018_CVPR}
Yuhua Chen, Wen Li, Christos Sakaridis, Dengxin Dai, and Luc Van~Gool.
\newblock Domain adaptive faster r-cnn for object detection in the wild.
\newblock In \emph{Proceedings of the IEEE Conference on Computer Vision and Pattern Recognition (CVPR)}, 2018.

\bibitem[Cordts et~al.(2016)Cordts, Omran, Ramos, Rehfeld, Enzweiler, Benenson, Franke, Roth, and Schiele]{7780719}
Marius Cordts, Mohamed Omran, Sebastian Ramos, Timo Rehfeld, Markus Enzweiler, Rodrigo Benenson, Uwe Franke, Stefan Roth, and Bernt Schiele.
\newblock The cityscapes dataset for semantic urban scene understanding.
\newblock In \emph{2016 IEEE Conference on Computer Vision and Pattern Recognition (CVPR)}, pages 3213--3223, 2016.

\bibitem[Cui et~al.(2024)Cui, Li, Zhang, Yan, Wang, Wang, Heng, and Li]{cui2024stochastic}
Yiming Cui, Liang Li, Jiehua Zhang, Chenggang Yan, Hongkui Wang, Shuai Wang, Jin Heng, and Wu Li.
\newblock Stochastic context consistency reasoning for domain adaptive object detection.
\newblock In \emph{ACM Multimedia 2024}, 2024.

\bibitem[Deng et~al.(2021)Deng, Li, Chen, and Duan]{Deng_2021_CVPR}
Jinhong Deng, Wen Li, Yuhua Chen, and Lixin Duan.
\newblock Unbiased mean teacher for cross-domain object detection.
\newblock In \emph{Proceedings of the IEEE/CVF Conference on Computer Vision and Pattern Recognition (CVPR)}, pages 4091--4101, 2021.

\bibitem[Deng et~al.(2023)Deng, Xu, Li, and Duan]{Deng_2023_CVPR}
Jinhong Deng, Dongli Xu, Wen Li, and Lixin Duan.
\newblock Harmonious teacher for cross-domain object detection.
\newblock In \emph{Proceedings of the IEEE/CVF Conference on Computer Vision and Pattern Recognition (CVPR)}, pages 23829--23838, 2023.

\bibitem[Feng et~al.(2024)Feng, Li, Gao, Huang, Zhang, Liu, and Wang]{pmlr-v235-feng24d}
Yongchao Feng, Shiwei Li, Yingjie Gao, Ziyue Huang, Yanan Zhang, Qingjie Liu, and Yunhong Wang.
\newblock {DSD}-{DA}: Distillation-based source debiasing for domain adaptive object detection.
\newblock In \emph{Proceedings of the 41st International Conference on Machine Learning}, pages 13225--13240. PMLR, 2024.

\bibitem[Ganin and Lempitsky(2015)]{pmlr-v37-ganin15}
Yaroslav Ganin and Victor Lempitsky.
\newblock Unsupervised domain adaptation by backpropagation.
\newblock In \emph{Proceedings of the 32nd International Conference on Machine Learning}, pages 1180--1189, Lille, France, 2015. PMLR.

\bibitem[Gonzalez and Woods(2018)]{gonzalez_digital_2018}
Rafael~C. Gonzalez and Richard~E. Woods.
\newblock \emph{Digital image processing}.
\newblock Pearson, New York, NY, 2018.

\bibitem[Gupta and Rao(1990)]{106875}
A. Gupta and K.R. Rao.
\newblock A fast recursive algorithm for the discrete sine transform.
\newblock \emph{IEEE Transactions on Acoustics, Speech, and Signal Processing}, 38\penalty0 (3):\penalty0 553--557, 1990.

\bibitem[He et~al.(2016)He, Zhang, Ren, and Sun]{He_2016_CVPR}
Kaiming He, Xiangyu Zhang, Shaoqing Ren, and Jian Sun.
\newblock Deep residual learning for image recognition.
\newblock In \emph{Proceedings of the IEEE Conference on Computer Vision and Pattern Recognition (CVPR)}, 2016.

\bibitem[He et~al.(2020)He, Fan, Wu, Xie, and Girshick]{He_2020_CVPR}
Kaiming He, Haoqi Fan, Yuxin Wu, Saining Xie, and Ross Girshick.
\newblock Momentum contrast for unsupervised visual representation learning.
\newblock In \emph{Proceedings of the IEEE/CVF Conference on Computer Vision and Pattern Recognition (CVPR)}, 2020.

\bibitem[He et~al.(2022)He, Wang, Wu, Wang, Li, Li, Gan, Wu, and Qiao]{He_2022_CVPR}
Mengzhe He, Yali Wang, Jiaxi Wu, Yiru Wang, Hanqing Li, Bo Li, Weihao Gan, Wei Wu, and Yu Qiao.
\newblock Cross domain object detection by target-perceived dual branch distillation.
\newblock In \emph{Proceedings of the IEEE/CVF Conference on Computer Vision and Pattern Recognition (CVPR)}, pages 9570--9580, 2022.

\bibitem[He and Zhang(2019)]{He_2019_ICCV}
Zhenwei He and Lei Zhang.
\newblock Multi-adversarial faster-rcnn for unrestricted object detection.
\newblock In \emph{Proceedings of the IEEE/CVF International Conference on Computer Vision (ICCV)}, 2019.

\bibitem[Howard et~al.(2017)Howard, Zhu, Chen, Kalenichenko, Wang, Weyand, Andreetto, and Adam]{Howard2017MobileNetsEC}
Andrew~G. Howard, Menglong Zhu, Bo Chen, Dmitry Kalenichenko, Weijun Wang, Tobias Weyand, Marco Andreetto, and Hartwig Adam.
\newblock Mobilenets: Efficient convolutional neural networks for mobile vision applications.
\newblock \emph{ArXiv}, abs/1704.04861, 2017.

\bibitem[Hoyer et~al.(2023)Hoyer, Dai, Wang, and Van~Gool]{10203852}
Lukas Hoyer, Dengxin Dai, Haoran Wang, and Luc Van~Gool.
\newblock Mic: Masked image consistency for context-enhanced domain adaptation.
\newblock In \emph{2023 IEEE/CVF Conference on Computer Vision and Pattern Recognition (CVPR)}, pages 11721--11732, 2023.

\bibitem[Huang et~al.(2021)Huang, Guan, Xiao, and Lu]{Huang_2021_CVPR}
Jiaxing Huang, Dayan Guan, Aoran Xiao, and Shijian Lu.
\newblock Fsdr: Frequency space domain randomization for domain generalization.
\newblock In \emph{Proceedings of the IEEE/CVF Conference on Computer Vision and Pattern Recognition (CVPR)}, pages 6891--6902, 2021.

\bibitem[Jiao et~al.(2023)Jiao, Yao, and Xu]{9935311}
Yifan Jiao, Hantao Yao, and Changsheng Xu.
\newblock Dual instance-consistent network for cross-domain object detection.
\newblock \emph{IEEE Transactions on Pattern Analysis and Machine Intelligence}, 45\penalty0 (6):\penalty0 7338--7352, 2023.

\bibitem[Johnson-Roberson et~al.(2016)Johnson-Roberson, Barto, Mehta, Sridhar, Rosaen, and Vasudevan]{JohnsonRoberson2016DrivingIT}
Matthew Johnson-Roberson, Charlie Barto, Rounak Mehta, Sharath~Nittur Sridhar, Karl Rosaen, and Ram Vasudevan.
\newblock Driving in the matrix: Can virtual worlds replace human-generated annotations for real world tasks?
\newblock \emph{2017 IEEE International Conference on Robotics and Automation (ICRA)}, pages 746--753, 2016.

\bibitem[Kennerley et~al.(2023)Kennerley, Wang, Veeravalli, and Tan]{Kennerley_2023_CVPR}
Mikhail Kennerley, Jian-Gang Wang, Bharadwaj Veeravalli, and Robby~T. Tan.
\newblock 2pcnet: Two-phase consistency training for day-to-night unsupervised domain adaptive object detection.
\newblock In \emph{Proceedings of the IEEE/CVF Conference on Computer Vision and Pattern Recognition (CVPR)}, pages 11484--11493, 2023.

\bibitem[Kennerley et~al.(2024)Kennerley, Wang, Veeravalli, and Tan]{10658175}
Mikhail Kennerley, Jian-Gang Wang, Bharadwaj Veeravalli, and Robby~T. Tan.
\newblock Cat: Exploiting inter-class dynamics for domain adaptive object detection.
\newblock In \emph{2024 IEEE/CVF Conference on Computer Vision and Pattern Recognition (CVPR)}, pages 16541--16550, 2024.

\bibitem[Lavoie et~al.(2025)Lavoie, Mahmoud, and Waslander]{Lavoie_2025_CVPR}
Marc-Antoine Lavoie, Anas Mahmoud, and Steven~L. Waslander.
\newblock Large self-supervised models bridge the gap in domain adaptive object detection.
\newblock In \emph{Proceedings of the IEEE/CVF Conference on Computer Vision and Pattern Recognition (CVPR)}, pages 4692--4702, 2025.

\bibitem[Li et~al.(2023{\natexlab{a}})Li, Zhang, Yao, Song, Hao, Zhao, Li, and Chen]{NEURIPS2023_0d18ab3b}
Haochen Li, Rui Zhang, Hantao Yao, Xinkai Song, Yifan Hao, Yongwei Zhao, Ling Li, and Yunji Chen.
\newblock Learning domain-aware detection head with prompt tuning.
\newblock In \emph{Advances in Neural Information Processing Systems}, pages 4248--4262. Curran Associates, Inc., 2023{\natexlab{a}}.

\bibitem[Li et~al.(2024)Li, Zhang, Yao, Zhang, Hao, Song, and Li]{10552660}
Haochen Li, Rui Zhang, Hantao Yao, Xin Zhang, Yifan Hao, Xinkai Song, and Ling Li.
\newblock React: Remainder adaptive compensation for domain adaptive object detection.
\newblock \emph{IEEE Transactions on Image Processing}, 33:\penalty0 3735--3748, 2024.

\bibitem[Li et~al.(2025)Li, Zhang, Yao, Zhang, Hao, Song, Peng, Zhao, Zhao, Wu, and Li]{Li_2025_CVPR}
Haochen Li, Rui Zhang, Hantao Yao, Xin Zhang, Yifan Hao, Xinkai Song, Shaohui Peng, Yongwei Zhao, Chen Zhao, Yanjun Wu, and Ling Li.
\newblock Seen-da: Semantic entropy guided domain-aware attention for domain adaptive object detection.
\newblock In \emph{Proceedings of the IEEE/CVF Conference on Computer Vision and Pattern Recognition (CVPR)}, pages 25465--25475, 2025.

\bibitem[Li et~al.(2022{\natexlab{a}})Li, Liu, Yao, and Yuan]{Li_Liu_Yao_Yuan_2022}
Wuyang Li, Xinyu Liu, Xiwen Yao, and Yixuan Yuan.
\newblock Scan: Cross domain object detection with semantic conditioned adaptation.
\newblock \emph{Proceedings of the AAAI Conference on Artificial Intelligence}, 36\penalty0 (2):\penalty0 1421--1428, 2022{\natexlab{a}}.

\bibitem[Li et~al.(2022{\natexlab{b}})Li, Liu, and Yuan]{Li_2022_CVPR}
Wuyang Li, Xinyu Liu, and Yixuan Yuan.
\newblock Sigma: Semantic-complete graph matching for domain adaptive object detection.
\newblock In \emph{Proceedings of the IEEE/CVF Conference on Computer Vision and Pattern Recognition (CVPR)}, pages 5291--5300, 2022{\natexlab{b}}.

\bibitem[Li et~al.(2023{\natexlab{b}})Li, Liu, and Yuan]{10012542}
Wuyang Li, Xinyu Liu, and Yixuan Yuan.
\newblock Sigma++: Improved semantic-complete graph matching for domain adaptive object detection.
\newblock \emph{IEEE Transactions on Pattern Analysis and Machine Intelligence}, 45\penalty0 (7):\penalty0 9022--9040, 2023{\natexlab{b}}.

\bibitem[Li et~al.(2022{\natexlab{c}})Li, Dai, Ma, Liu, Chen, Wu, He, Kitani, and Vajda]{li2022cross}
Yu-Jhe Li, Xiaoliang Dai, Chih-Yao Ma, Yen-Cheng Liu, Kan Chen, Bichen Wu, Zijian He, Kris Kitani, and Peter Vajda.
\newblock Cross-domain adaptive teacher for object detection.
\newblock In \emph{IEEE Conference on Computer Vision and Pattern Recognition (CVPR)}, 2022{\natexlab{c}}.

\bibitem[Lin et~al.(2023)Lin, Zhang, Huang, Lu, Lan, Chu, You, Wang, Liu, Parulkar, Navkal, and Chen]{Lin_2023_CVPR}
Shiqi Lin, Zhizheng Zhang, Zhipeng Huang, Yan Lu, Cuiling Lan, Peng Chu, Quanzeng You, Jiang Wang, Zicheng Liu, Amey Parulkar, Viraj Navkal, and Zhibo Chen.
\newblock Deep frequency filtering for domain generalization.
\newblock In \emph{Proceedings of the IEEE/CVF Conference on Computer Vision and Pattern Recognition (CVPR)}, pages 11797--11807, 2023.

\bibitem[Lin et~al.(2014)Lin, Maire, Belongie, Hays, Perona, Ramanan, Doll{\'a}r, and Zitnick]{Lin2014MicrosoftCC}
Tsung-Yi Lin, Michael Maire, Serge~J. Belongie, James Hays, Pietro Perona, Deva Ramanan, Piotr Doll{\'a}r, and C.~Lawrence Zitnick.
\newblock Microsoft coco: Common objects in context.
\newblock In \emph{European Conference on Computer Vision}, 2014.

\bibitem[Liu et~al.(2022)Liu, Li, Yang, Li, and Yuan]{Liu_2022_CVPR}
Xinyu Liu, Wuyang Li, Qiushi Yang, Baopu Li, and Yixuan Yuan.
\newblock Towards robust adaptive object detection under noisy annotations.
\newblock In \emph{Proceedings of the IEEE/CVF Conference on Computer Vision and Pattern Recognition (CVPR)}, pages 14207--14216, 2022.

\bibitem[Liu et~al.(2023)Liu, Wang, Huang, Wang, and Xu]{10204361}
Yabo Liu, Jinghua Wang, Chao Huang, Yaowei Wang, and Yong Xu.
\newblock Cigar: Cross-modality graph reasoning for domain adaptive object detection.
\newblock In \emph{2023 IEEE/CVF Conference on Computer Vision and Pattern Recognition (CVPR)}, pages 23776--23786, 2023.

\bibitem[Oquab et~al.(2024)Oquab, Darcet, Moutakanni, Vo, Szafraniec, Khalidov, Fernandez, HAZIZA, Massa, El-Nouby, Assran, Ballas, Galuba, Howes, Huang, Li, Misra, Rabbat, Sharma, Synnaeve, Xu, Jegou, Mairal, Labatut, Joulin, and Bojanowski]{oquab2024dinov}
Maxime Oquab, Timoth{\'e}e Darcet, Th{\'e}o Moutakanni, Huy~V. Vo, Marc Szafraniec, Vasil Khalidov, Pierre Fernandez, Daniel HAZIZA, Francisco Massa, Alaaeldin El-Nouby, Mido Assran, Nicolas Ballas, Wojciech Galuba, Russell Howes, Po-Yao Huang, Shang-Wen Li, Ishan Misra, Michael Rabbat, Vasu Sharma, Gabriel Synnaeve, Hu Xu, Herve Jegou, Julien Mairal, Patrick Labatut, Armand Joulin, and Piotr Bojanowski.
\newblock {DINO}v2: Learning robust visual features without supervision.
\newblock \emph{Transactions on Machine Learning Research}, 2024.
\newblock Featured Certification.

\bibitem[Piao et~al.(2024)Piao, Tang, and Zhao]{10154573}
Zhengquan Piao, Linbo Tang, and Baojun Zhao.
\newblock Unsupervised domain-adaptive object detection via localization regression alignment.
\newblock \emph{IEEE Transactions on Neural Networks and Learning Systems}, 35\penalty0 (11):\penalty0 15170--15181, 2024.

\bibitem[Radford et~al.(2021)Radford, Kim, Hallacy, Ramesh, Goh, Agarwal, Sastry, Askell, Mishkin, Clark, Krueger, and Sutskever]{radford2021learning}
Alec Radford, Jong~Wook Kim, Chris Hallacy, Aditya Ramesh, Gabriel Goh, Sandhini Agarwal, Girish Sastry, Amanda Askell, Pamela Mishkin, Jack Clark, Gretchen Krueger, and Ilya Sutskever.
\newblock Learning transferable visual models from natural language supervision.
\newblock In \emph{Proceedings of the 38th International Conference on Machine Learning (ICML)}, pages 8748--8763. PMLR, 2021.

\bibitem[Ren et~al.(2015)Ren, He, Girshick, and Sun]{NIPS2015_14bfa6bb}
Shaoqing Ren, Kaiming He, Ross Girshick, and Jian Sun.
\newblock Faster r-cnn: Towards real-time object detection with region proposal networks.
\newblock In \emph{Advances in Neural Information Processing Systems}. Curran Associates, Inc., 2015.

\bibitem[Saito et~al.(2019)Saito, Ushiku, Harada, and Saenko]{Saito_2019_CVPR}
Kuniaki Saito, Yoshitaka Ushiku, Tatsuya Harada, and Kate Saenko.
\newblock Strong-weak distribution alignment for adaptive object detection.
\newblock In \emph{Proceedings of the IEEE/CVF Conference on Computer Vision and Pattern Recognition (CVPR)}, 2019.

\bibitem[Sakaridis et~al.(2017)Sakaridis, Dai, and Gool]{Sakaridis2017SemanticFS}
Christos Sakaridis, Dengxin Dai, and Luc~Van Gool.
\newblock Semantic foggy scene understanding with synthetic data.
\newblock \emph{International Journal of Computer Vision}, 126:\penalty0 973 -- 992, 2017.

\bibitem[Sakaridis et~al.(2021)Sakaridis, Dai, and Van~Gool]{Sakaridis_2021_ICCV}
Christos Sakaridis, Dengxin Dai, and Luc Van~Gool.
\newblock Acdc: The adverse conditions dataset with correspondences for semantic driving scene understanding.
\newblock In \emph{Proceedings of the IEEE/CVF International Conference on Computer Vision (ICCV)}, pages 10765--10775, 2021.

\bibitem[Sakaridis et~al.(2025)Sakaridis, Bruggemann, Yu, and Van~Gool]{10840277}
Christos Sakaridis, David Bruggemann, Fisher Yu, and Luc Van~Gool.
\newblock Condition-invariant semantic segmentation.
\newblock \emph{IEEE Transactions on Pattern Analysis and Machine Intelligence}, 47\penalty0 (4):\penalty0 3111--3125, 2025.

\bibitem[Su et~al.(2020)Su, Wang, Zeng, Tang, Chen, Qiu, and Wang]{su2020adapting}
Peng Su, Kun Wang, Xingyu Zeng, Shixiang Tang, Dapeng Chen, Di Qiu, and Xiaogang Wang.
\newblock Adapting object detectors with conditional domain normalization.
\newblock In \emph{Proceedings of the European Conference on Computer Vision (ECCV)}, pages 403--419. Springer, 2020.

\bibitem[Tarvainen and Valpola(2017)]{NIPS2017_68053af2}
Antti Tarvainen and Harri Valpola.
\newblock Mean teachers are better role models: Weight-averaged consistency targets improve semi-supervised deep learning results.
\newblock In \emph{Advances in Neural Information Processing Systems}. Curran Associates, Inc., 2017.

\bibitem[Wang et~al.(2020)Wang, Wu, Huang, and Xing]{Wang_2020_CVPR}
Haohan Wang, Xindi Wu, Zeyi Huang, and Eric~P. Xing.
\newblock High-frequency component helps explain the generalization of convolutional neural networks.
\newblock In \emph{Proceedings of the IEEE/CVF Conference on Computer Vision and Pattern Recognition (CVPR)}, 2020.

\bibitem[Xu et~al.(2020)Xu, Zhao, Jin, and Wei]{Xu_2020_CVPR}
Chang-Dong Xu, Xing-Ran Zhao, Xin Jin, and Xiu-Shen Wei.
\newblock Exploring categorical regularization for domain adaptive object detection.
\newblock In \emph{Proceedings of the IEEE/CVF Conference on Computer Vision and Pattern Recognition (CVPR)}, 2020.

\bibitem[Yang and Soatto(2020)]{Yang_2020_CVPR}
Yanchao Yang and Stefano Soatto.
\newblock Fda: Fourier domain adaptation for semantic segmentation.
\newblock In \emph{Proceedings of the IEEE/CVF Conference on Computer Vision and Pattern Recognition (CVPR)}, 2020.

\bibitem[Yoo et~al.(2022)Yoo, Chung, and Kwak]{10.1007/978-3-031-19827-4_40}
Jayeon Yoo, Inseop Chung, and Nojun Kwak.
\newblock Unsupervised domain adaptation for one-stage object detector using offsets to bounding box.
\newblock In \emph{Computer Vision -- ECCV 2022}, pages 691--708, Cham, 2022. Springer Nature Switzerland.

\bibitem[Yu et~al.(2020)Yu, Chen, Wang, Xian, Chen, Liu, Madhavan, and Darrell]{9156329}
Fisher Yu, Haofeng Chen, Xin Wang, Wenqi Xian, Yingying Chen, Fangchen Liu, Vashisht Madhavan, and Trevor Darrell.
\newblock Bdd100k: A diverse driving dataset for heterogeneous multitask learning.
\newblock In \emph{2020 IEEE/CVF Conference on Computer Vision and Pattern Recognition (CVPR)}, pages 2633--2642, 2020.

\bibitem[Zamir et~al.(2022)Zamir, Arora, Khan, Hayat, Khan, and Yang]{Zamir_2022_CVPR}
Syed~Waqas Zamir, Aditya Arora, Salman Khan, Munawar Hayat, Fahad~Shahbaz Khan, and Ming-Hsuan Yang.
\newblock Restormer: Efficient transformer for high-resolution image restoration.
\newblock In \emph{Proceedings of the IEEE/CVF Conference on Computer Vision and Pattern Recognition (CVPR)}, pages 5728--5739, 2022.

\bibitem[Zhong et~al.(2022)Zhong, Yang, Zhang, Li, Codella, Li, Zhou, Dai, Yuan, Li, and Gao]{Zhong_2022_CVPR}
Yiwu Zhong, Jianwei Yang, Pengchuan Zhang, Chunyuan Li, Noel Codella, Liunian~Harold Li, Luowei Zhou, Xiyang Dai, Lu Yuan, Yin Li, and Jianfeng Gao.
\newblock Regionclip: Region-based language-image pretraining.
\newblock In \emph{Proceedings of the IEEE/CVF Conference on Computer Vision and Pattern Recognition (CVPR)}, pages 16793--16803, 2022.

\bibitem[Zhou et~al.(2023)Zhou, Fan, Luo, and Zhang]{10376923}
Wenzhang Zhou, Heng Fan, Tiejian Luo, and Libo Zhang.
\newblock Unsupervised domain adaptive detection with network stability analysis.
\newblock In \emph{2023 IEEE/CVF International Conference on Computer Vision (ICCV)}, pages 6963--6972, 2023.

\bibitem[Zhu et~al.(2017)Zhu, Park, Isola, and Efros]{CycleGAN2017}
Jun-Yan Zhu, Taesung Park, Phillip Isola, and Alexei~A Efros.
\newblock Unpaired image-to-image translation using cycle-consistent adversarial networks.
\newblock In \emph{Computer Vision (ICCV), 2017 IEEE International Conference on}, 2017.

\bibitem[Zhu et~al.(2019)Zhu, Pang, Yang, Shi, and Lin]{Zhu_2019_CVPR}
Xinge Zhu, Jiangmiao Pang, Ceyuan Yang, Jianping Shi, and Dahua Lin.
\newblock Adapting object detectors via selective cross-domain alignment.
\newblock In \emph{Proceedings of the IEEE/CVF Conference on Computer Vision and Pattern Recognition (CVPR)}, 2019.

\end{thebibliography}
}

\end{document}